\documentclass[conference]{IEEEtran}
\usepackage{etoolbox}
\usepackage{amsmath}
\usepackage{cite}
\usepackage{algorithmic}
\usepackage{graphicx}
\usepackage{textcomp}
\usepackage{enumitem}
\usepackage{calc}
\usepackage{multirow}

\graphicspath{{../}}

\def\BibTeX{{\rm B\kern-.05em{\sc i\kern-.025em b}\kern-.08em
    T\kern-.1667em\lower.7ex\hbox{E}\kern-.125emX}}

\AtBeginEnvironment{align}{\setcounter{equation}{0}}
\AtBeginEnvironment{equation}{\setcounter{equation}{0}}

\begin{document}

%\xdef\xfigwd{\the\wd\figbox}
    
%\history{Date of publication xxxx 00, 0000, date of current version xxxx 00, 0000.}
%\doi{10.1109/ACCESS.2017.DOI}

\title{Automatic Design of Artificial Neural \\Networks for Gamma-Ray Detection}

%\author{Filipe Assun\c{c}\~ao\authorrefmark{1}, Jo\~ao Correia\authorrefmark{1}, R\'uben Concei\c{c}\~ao\authorrefmark{2}, M\'ario Pimenta\authorrefmark{2}, Bernardo Tom\'e\authorrefmark{2}, Nuno Louren\c{c}o\authorrefmark{1}, Penousal Machado\authorrefmark{1}}
%\address[1]{CISUC, Department of Informatics Engineering, University of Coimbra, Portugal}
%\address[2]{LIP/IST, Lisboa, Portugal} 
%\tfootnote{This work was partially funded by Funda\c{c}\~ao para a Ci\^encia e Tecnologia (FCT), Portugal, under the PhD grant SFRH/BD/114865/2016.}

%\markboth
%{F. Assun\c{c}\~ao  \headeretal: Automatic Design of Artificial Neural Networks for Gamma-Ray Detection}
%{F. Assun\c{c}\~ao  \headeretal: Automatic Design of Artificial Neural Networks for Gamma-Ray Detection}

%\corresp{Corresponding author: Filipe Assun\c{c}\~ao (e-mail: fga@dei.uc.pt).}

\author{\IEEEauthorblockN{Filipe Assun\c{c}\~ao\IEEEauthorrefmark{1},
Jo\~ao Correia\IEEEauthorrefmark{1}, R\'uben Concei\c{c}\~ao\IEEEauthorrefmark{2},
M\'ario Pimenta\IEEEauthorrefmark{2},\\ Bernardo Tom\'e\IEEEauthorrefmark{2},
Nuno Louren\c{c}o\IEEEauthorrefmark{1}, Penousal Machado\IEEEauthorrefmark{1}}

\IEEEauthorblockA{\IEEEauthorrefmark{1}CISUC, Department of Informatics Engineering, University of Coimbra, Portugal,\\ \IEEEauthorrefmark{2}LIP/IST, Lisboa, Portugal\\
E-mails: \{fga,jncor\}@dei.uc.pt, \{ruben,pimenta,bernardo\}@lip.pt, \{naml,machado\}@dei.uc.pt}

}

% \author{\IEEEauthorblockN{Filipe Assun\c{c}\~ao}
% \IEEEauthorblockA{CISUC \\
% University of Coimbra\\
% Coimbra, Portugal \\
% fga@dei.uc.pt}
% \and
% \IEEEauthorblockN{Jo\~ao Correia}
% \IEEEauthorblockA{CISUC\\
% University of Coimbra\\
% Coimbra, Portugal \\
% jncor@dei.uc.pt}
% \and
% \IEEEauthorblockN{R\'uben Concei\c{c}\~ao}
% \IEEEauthorblockA{LIP/IST \\
% Lisboa, Portugal\\
% ruben@lip.pt}
% \and
% \IEEEauthorblockN{M\'ario Pimenta}
% \IEEEauthorblockA{LIP/IST \\
% Lisboa, Portugal\\
% pimenta@lip.pt}
% \and
% \IEEEauthorblockN{Bernardo Tom\'e}
% \IEEEauthorblockA{LIP/IST \\
% Lisboa, Portugal\\
% bernardo@lip.pt}
% \and
% \IEEEauthorblockN{Nuno Louren\c{co}}
% \IEEEauthorblockA{CISUC\\
% University of Coimbra\\
% Coimbra, Portugal \\
% naml@dei.uc.pt}
% \and 
% \IEEEauthorblockN{Penousal Machado}
% \IEEEauthorblockA{CISUC\\
% University of Coimbra\\
% Coimbra, Portugal \\
% machado@dei.uc.pt}
% }

%\author{Filipe Assun\c{c}\~ao, Jo\~ao Correia, R\'uben Concei\c{c}\~ao, M\'ario Pimenta, Bernardo Tom\'e, Nuno Louren\c{c}o, Penousal Machado}

\maketitle

\begin{abstract}
The goal of this work is to investigate the possibility of improving current gamma/hadron discrimination based on their shower patterns recorded on the ground. To this end we propose the use of Convolutional Neural Networks (CNNs) for their ability to distinguish patterns based on automatically designed features. In order to promote the creation of CNNs that properly uncover the hidden patterns in the data, and at same time avoid the burden of hand-crafting the topology and learning hyper-parameters we resort to NeuroEvolution; in particular we use Fast-DENSER++, a variant of Deep Evolutionary Network Structured Representation. The results show that the best CNN generated by Fast-DENSER++ improves by a factor of 2 when compared with the results reported by classic statistical approaches. Additionally, we experiment ensembling the 10 best generated CNNs, one from each of the evolutionary runs; the ensemble leads to an improvement by a factor of 2.3. These results show that it is possible to improve the gamma/hadron discrimination based on CNNs that are automatically generated and are trained with instances of the ground impact patterns.
\end{abstract}

%\begin{keywords}
%Artificial Neural Networks, Evolutionary Computation, Gamma-Ray Detection
%\end{keywords}

%\titlepgskip=-15pt

%\section{Detection of Gamma Rays}
%\label{sec:problem_description}

\section{Introduction}
\label{sec:introduction}

High-energy gamma-rays constitute one of the best probes to investigate extreme phenomena in the Universe, such gamma-rays arising from fast rotating neutron stars or supermassive black holes.
The detection of this kind of astrophysical radiation, whose energies span from 10 GeV up to 100 TeV, can be done at lower energies by satellite bourne detectors. However, above a few hundreds GeV, the flux becomes too small, and only ground-based experiments can measure indirectly gamma-rays. These experiments take advantage of the electromagnetic cascade that is produced by the interaction of gamma-rays with Earth's atmosphere to infer the direction and energy of the primary gamma-ray.
If the energy of the gamma-ray is sufficiently high and the detection of the secondary shower particles is done at high altitude, then it is possible to survey large portions of the sky and be sensitive to transient phenomena.
The observation of high-energy gamma-rays with ground-arrays, although effective, comes with a cost: one has to deal with the huge background of cosmic rays that bombard the Earth continuously.
To select gamma-rays out of the hadronic background one can explore the characteristics of the shower development. Contrary to pure electromagnetic showers, hadron induced showers produce high transverse momentum particles which lead to the transverse broadening of the shower and the creation of clusters.
Experimentally, the above features can be explored by measuring the steepness and bumpiness of the lateral distribution of particles at the ground with respect to the shower core position or by measuring the relative amount of signal (number of particles) at large distances from the shower core.
However, the patterns of the secondary particles at the ground remain to be explored, although some studies have shown that this might have some gamma/hadron discrimination power.
In this manuscript, we intend to explore the difference in the patterns at the ground, between gamma and proton induced showers, recurring to Artificial Neural Networks (ANNs). We compare the performance of ANNs to the performance of classic statistical approaches that resort to human-extracted features. To overcome the difficulty associated to the design of ANNs we use NeuroEvolution to automate the choice for the topology and learning strategy (Section~\ref{sec:neuroevolution}); in particular we use Fast Deep Evolutionary Network Structure Representation ++ (F-DENSER++), detailed in Section~\ref{sec:denser}. The results (Section~\ref{sec:evolution}) show that the performance of the fittest network generated by F-DENSER++ surpasses the performance of classic statistical approaches. The gains in performance represent an improvement by a factor of 2.3; this indicates that with the same grid of sensors we can perform twice better than other methods; on the other hand it can lead to investment saving because a smaller grid of detectors can be used.

\section{Gamma and proton simulation}
\label{sec:simulation}

The above proposed investigations were done using gamma and proton (hadron) simulations, generated with CORSIKA~\cite{corsika}, and an experiment layout as described in~\cite{LATTES}. The detectors have been simulated with the Geant4 toolkit~\cite{geant4} and the recorded signals have been used to reconstruct the main shower characteristics (energy, direction, primary) so that the sensitivity of this experiment to gamma-ray sources could be evaluated realistically.
The detector unit is composed of small water Cherenkov detectors, which maximizes the trigger efficiency, and segmented resistive plate chambers, which have a good time resolution providing in this way a good shower geometry reconstruction. This detector concept was chosen to lower the energy threshold of previous experiments and bridge the energy gap between satellite-bourne and present ground-based experiments.

The main aim of this work is to prove that the analysis of the pattern at the ground can be used to improve current gamma/hadron discrimination techniques. As such, we have opted to use only for the present study the information of the water-Cherenkov detectors (WCDs). Moreover, only showers reconstructed with energies between $1$ and $1.7$\,TeV were used.
Secondary shower particles that hit the WCD will produce light that can be recorded by photomultipliers mounted sideways. As such, for each shower event, a WCD station provides the following information: its position (x and y coordinates of the center of the WCD), and the recorded signal (approximately proportional to the number of particles in it). It is only this information that shall be used to distinguish gamma from hadron induced showers.

In~\cite{LATTES}, it was demonstrated that this detector concept can perform the usual gamma/hadron discrimination. Two discrimination variables based solely in the WCD information where built: \emph{Compactness} and \emph{S40}. The former explores the information in the shower lateral distribution function (LDF), in particular, the steepness and bumpiness. This is done comparing the shower event LDF to a reference gamma LDF, built from the average of many gamma showers. The variable S40 is used to identify particle clusters away from the shower core. This is achieved by computing, for stations above 40 meters away of the reconstructed shower core, the ratio between the signal of the hottest station and the total signal. Although there is some level of correlation between the two variables, they carry independent information. To further explore the combined discrimination power of \emph{Compactness} and \emph{S40}, a linear discriminant analysis is used, henceforth referred simply as \emph{Fisher}. It is worth to mention that although the above quantities are certainly exploring the shower pattern at the ground, these classical statistics analyses cannot fully extract all the information due to the stochastic nature of the shower, forcing the use of non-parametric cuts.

\begin{figure}[t!]
    \scriptsize
    \begin{align}
        {<}\text{fully-connected}{>} ::= & \, \text{layer:fc} \, {<}\text{activation}{>} \\
                    & \, \text{[num-units,int,1,128,2048} \, {<}\text{bias}{>} \\
        {<}\text{dropout}{>} ::= & \text{layer:dropput} \, \text{[rate,float,1,0,0.7]} \\
        {<}\text{activation}{>} ::= & \, \text{act:linear} \, | \, \text{act:relu} \, | \, \text{act:sigmoid}\\
        {<}\text{bias}{>} ::= & \, \text{bias:True} \, | \, \text{bias:False}\\
        {<}\text{softmax}{>} ::= & \, \text{layer:fc} \, \text{act:softmax} \, \text{num-units:10} \, \text{bias:True}\\
        {<}\text{learning}{>} ::= & \, {<}\text{bp}{>} \, \text{[batch\_size,int,1,50,500]} \\
                   & \, | \, {<}\text{rmsprop}{>} \, \text{[batch\_size,int,1,50,500]}  \\
                   & \, | \, {<}\text{adam}{>} \, \text{[batch\_size,int,1,50,500]} \\
         {<}\text{bp}{>} ::= & \, \text{learning:gradient-descent} \, \text{[lr,float,1,0.0001,0.1]}  \\
                   & \, \text{[momentum,float,1,0.68,0.99]} \\
                   & \, \text{[decay,float,1,0.000001,0.001]} \, {<}\text{nesterov}{>} \\
         {<}\text{nesterov}{>} ::= & \, \text{nesterov:True} \, | \, \text{nesterov:False} \\
         {<}\text{adam}{>} ::= & \, \text{learning:adam} \, \text{[lr,float,1,0.0001,0.1]} \\
                               & \, \text{[beta1,float,1,0.5,1]} \, \text{[beta2,float,1,0.5,1]} \\
                               & \, \text{[decay,float,1,0.000001,0.001]} \\
         {<}\text{rmsprop}{>} ::= & \, \text{learning:rmsprop} \, \text{[lr,float,1,0.0001,0.1]} \\
                   & \, \text{[rho,float,1,0.5,1]} \, \text{[decay,float,1,0.000001,0.001]} 
    \end{align}

    \small
    \caption{Example of a grammar for encoding fully-connected networks.}
    \label{fig:grammar_example}
\end{figure}

\section{NeuroEvolution}
\label{sec:neuroevolution}

NeuroEvolution (NE)~\cite{floreano2008neuroevolution} refers to the set of methods that apply Evolutionary Computation (EC) to automatically optimise ANNs. There are several NE approaches, which are often grouped according to the target of evolution. For example, Si et al.~\cite{7055036}, David and Greental~\cite{DBLP:conf/gecco/DavidG14}, and Morse and Stanley~\cite{DBLP:conf/gecco/MorseS16} optimise the \emph{synaptic weights}, Shabash et al.~\cite{shabash2018evonn} search for the weights and activation functions, and Radi and Poli~\cite{radi1998discovery} evolve neural network \emph{learning rules}. Differently, Soltanian et al.~\cite{6599788}, Suganuma et al.~\cite{DBLP:conf/gecco/SuganumaSN17}, and Fernando et al.~\cite{DBLP:journals/corr/FernandoBBZHRPW17} search only the topology.

\begin{figure*}[t!]
    \centering
    \includegraphics[width=.7\textwidth]{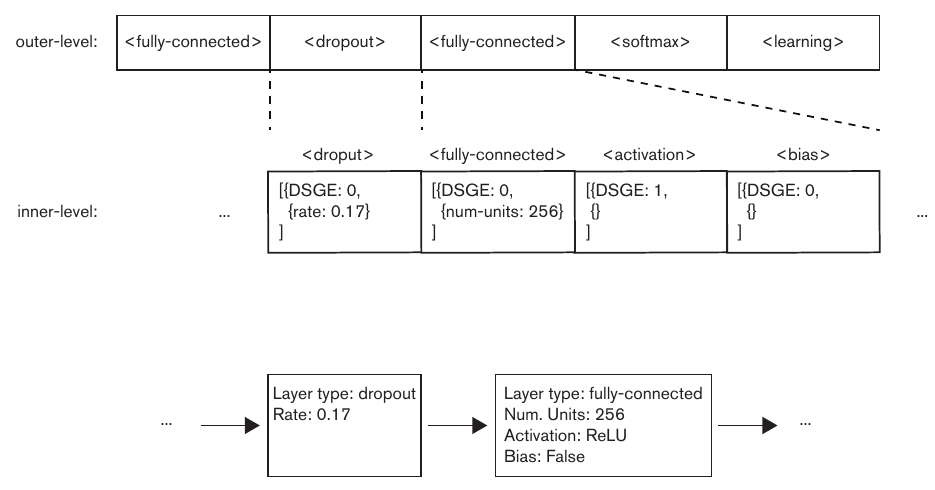}
    \caption{Example of a DENSER's genotype (top), and corresponding phenotype (bottom). The example is based on the outer-level structure [((fully-connected, dropout), 1, 10), (softmax, 1, 1), (learning, 1, 1)], and on the grammar of Figure~\ref{fig:grammar_example}.}
    \label{fig:genotype_phenotype_example}
\end{figure*}

The separate optimisation of either the learning strategy or the topology has proven successful. On the one hand, NE has shown to be competitive with standard (non-evolutionary) learning algorithms~\cite{DBLP:conf/ijcai/MontanaD89,DBLP:conf/gecco/MorseS16}, and does not require the activation functions to be differentiable. On the other hand, when optimising the structure of the network, the evolutionary results match (and even surpass) the ones attained by grid or random search, given less computational time~\cite{DBLP:conf/gecco/LorenzoNKRP17}. Nonetheless, Turner and Miller~\cite{DBLP:conf/sgai/TurnerM13} state that ``the choice of topology has a dramatic impact on the effectiveness of NE when only evolving weights; an issue not faced when manipulating both weights and topology'', and therefore it is beneficial to evolve the topology and weights simultaneously. Examples of methods that simultaneously search for the best weights and topology are ANNA Eleonora\cite{DBLP:journals/tnn/Maniezzo94}, NeuroEvolution of Augmenting Topologies (NEAT)~\cite{stanley2002evolving}, or Cartesian Genetic Programming Artificial Neural Networks (CGPANN)~\cite{DBLP:conf/gecco/TurnerM13}.

The previous methods work well on the optimisation of the weights and topology of small scale networks, i.e., ANNs with few neurons; however optimising hundreds or thousands of weights, and the topology of the network simultaneously is hard. That is the reason why the vast majority of the approaches that focus on the optimisation of deep networks~\cite{DBLP:journals/corr/MiikkulainenLMR17,assuncao2019denser,DBLP:journals/corr/abs-1710-10741} optimise the topology (e.g., number, type, and sequencing of layers), and the learning hyper-parameters rather than the weights, i.e., the methods focus on the optimisation of which learning algorithm to train the network (e.g., Backpropagation, or Adam), and its hyper-parameters (e.g., learning rate, or momentum).

One of the main drawbacks of NE concerns the time required for evaluating the population of candidate solutions. NE is on Evolutionary Computation, and thus a population of candidate solutions is evaluated throughout a (usually large) number of generations. To evaluate each candidate solution when the weights are not directly evolved we need to train the network, and when using large datasets the training process is time consuming. Therefore, the networks are often trained for a fixed (low) number of epochs (e.g., 8-10 epochs). To overcome the burden of evolution we can use clusters of Graphic Processing Units (GPUs) (e.g., Amazon AWS, or Google Cloud)~\cite{liang2019evolutionary}, evaluate the candidate solutions in a limited amount of data instances~\cite{DBLP:conf/gecco/MorseS16}, or train for a fixed amount of epochs/time and let evolution resume the training in a subsequent generation by loading the previous weights~\cite{DBLP:journals/tnn/YaoL97}.

In the current work we use a variant of Deep Evolutionary Network Structured Representation (DENSER)~\cite{assuncao2019denser} to search for Convolutional Neural Networks (CNNs) to distinguish between gamma radiations and protons. DENSER, and the reasons for selecting it are detailed in Section~\ref{sec:denser}.

\section{Deep Evolutionary Network Structured Representation}
\label{sec:denser}

Deep Evolutionary Network Structured Representation (DENSER)~\cite{assuncao2019denser}, is a general-purpose grammar-based NeuroEvolution (NE) approach. It has successfully been applied in object detection tasks, and all the user inputs are defined in a human-readable format, and thus the framework is easy to adapt to different domains and network structures.

In DENSER, the individuals are encoded using a two level representation: (i) the outer-level represents the macro-structure of the network, i.e., the sequence of evolutionary units\footnote{In DENSER the evolutionary units correspond to all aspects of the network that are to be optimised, e.g., layers and learning strategy, but can also include data pre-processing and data-augmentation blocks.}; and (ii) the inner-level keeps the parameters associated to the outer-level evolutionary unit. Whilst the outer-level is parameterised by the user-definition of a outer-level structure, the inner-level is parameterised by means of a Context-Free-Grammar (CFG). For example, for encoding a fully-connected network, with fully-connected and dropout layers the following outer-level structure can be defined: [((fully-connected, dropout), 1, 10), (softmax, 1, 1), (learning, 1, 1)]\footnote{The outer-level structure defines the network sequencing using the following format: [(production-rules, min\_evo\_units, max\_evo\_units), ...]}: that is, the network structure is composed by between 1 and 10 fully-connected and/or dropout evolutionary units, 1 softmax evolutionary unit, and 1 learning evolutionary unit. The outer-level structure production-rules require a one-to-one mapping to the grammar that is used for the inner-level. Figure~\ref{fig:grammar_example} encodes an example of a grammar; there is a production rule for fully-connected, dropout, softmax, and learning. The grammar encodes the parameterisation required for each of the parameters of the evolutionary units; the parameters can be of one of the following types: integer, float or closed choice, and the parameter block has the following format [variable-name, variable-type, num\_values, min\_value, max\_value].

\begin{figure*}[t!]
    \centering
    \includegraphics[width=0.9\textwidth]{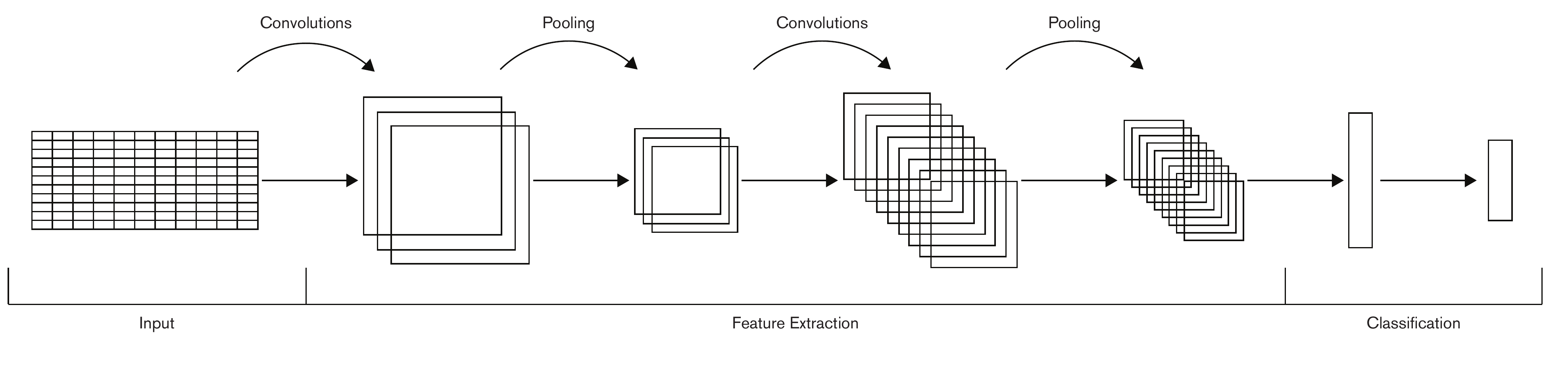}
    \caption{Topology of a Convolutional Neural Network.}
    \label{fig:cnns}
\end{figure*}

The evolutionary engine of the inner-level of DENSER is based on Dynamic Structured Grammatical Evolution (DSGE): a variant of Grammatical Evolution (GE)~\cite{o2001grammatical} that solves its redundancy, and locality issues; there is a one-to-one mapping between the expansion possibilities and the production rules, and the genotype grows as needed, meaning that there are no non-coding parts in the genotype. For more details on DSGE the reader should refer to~\cite{assunccao2017towards,lourencco2018structured}. 

An example of a genotype and phenotype of an individual using the outer-level-structure [((fully-connected, dropout), 1, 10), (softmax, 1, 1), (learning, 1, 1)], and the grammar of Figure~\ref{fig:grammar_example} is shown in Figure~\ref{fig:genotype_phenotype_example}. The individual of the example is an ANN with 4 layers (2 fully-connected, 1 dropout, and 1 softmax) and the learning strategy. The inner-level representation follows the standard of DSGE, where the DSGE integer represents the expansion possibility; e.g., from Figure~\ref{fig:grammar_example} we know that the activation non-terminal symbol has 3 expansion possibilities (linear, relu, or sigmoid), and therefore ``DSGE: 1'' on the example implies that we select the relu expansion (as evidenced on the phenotype). The inner-level genotype and the phenotype focus without loss of generality on two evolutionary units.

To promote evolution DENSER introduces genetic operators specifically tailored for the manipulation of ANNs. The mutations enable the addition, duplication\footnote{Whilst the addition creates a new evolutionary unit, at random, the duplication performs a copy by reference, i.e., if during evolution any of the copies parameters' is changed all copies are affected.}, or removal of evolutionary units (at the outer-level), and the perturbation of any of the parameters and expansion possibilities (at the inner-level). The crossover swaps evolutionary units.  

To assess the fitness of the individuals they are evaluated using either a fixed learning strategy (in case only the topology is the target of evolution), or the learning policy that constitutes an evolutionary unit (as in the grammar of Figure~\ref{fig:grammar_example}). The candidate solutions in DENSER are trained for a limited number of epochs (fixed to 10).

The following sub-sections detail two cumulative variants of DENSER: Fast-DENSER (Section~\ref{sec:fast_denser}), and Fast-DENSER++ (Section~\ref{sec:fast_denser_plus}). These variants solve issues of the standard DENSER version, that are pointed out next. 

\subsection{Fast-DENSER}
\label{sec:fast_denser}

Evolution in the standard DENSER implementation is carried out as typically in a Genetic Algorithm (GA), i.e., a population of individuals (often of size 100 or more) is evolved throughout a large number of generations. This requires a great number of evaluations, and thus slows down evolution. 

As the name suggests the end-goal of Fast-DENSER (F-DENSER)~\cite{assunccao2019fast} is to speedup evolution. To accomplish that F-DENSER replaces the GA evolutionary procedure by a (1+$\lambda$)-Evolutionary Strategy (ES); therefore, in each generation only 1+$\lambda$ individuals are evaluated. In the conducted experiments the authors compare DENSER (with a population size of 100 individuals), and F-DENSER (with $\lambda$=4) on the evolution of CNNs; therefore, whilst in DENSER in each generation 100 individuals are evaluated, in F-DENSER only 5 individuals are evaluated. The results demonstrate that the performance of DENSER and F-DENSER is the same, but F-DENSER takes, on average, 20x less time to generate the best solutions.

Another difference between F-DENSER and DENSER lies on the evaluation stop criteria. Instead of training each individual for a fixed number of 10 epochs, F-DENSER also investigates the evaluation of the individuals up to a maximum GPU time, i.e., all individuals are granted access the same computational resources. The training for a maximum granted GPU time makes the assessment of the learning strategy more adequate as more or less epochs can be performed depending on the network requirements. 

\subsection{Fast-DENSER++}
\label{sec:fast_denser_plus}

Despite the speedup of F-DENSER over DENSER, the method is not able to generate networks that are ready for deployment right-off evolution, i.e., during evolution the models are evaluated for a fixed number of epochs, or up to a maximum granted GPU time, but that does not guarantee that further training time does not increase the performance of the network. 

%TODO: fix ref
Fast-DENSER++~\cite{assuncao2019fdenser++} (F-DENSER++) builds on top of F-DENSER by introducing a new mutation operator that modifies the maximum training time that is granted to each individual. The rationale is to increase the training as the networks grow, i.e., during the initial generations the networks tend to be simple and therefore require less evaluation time; as time proceeds, the networks become more complex and may benefit from longer trains. 

In the current paper, we conduct the experiments with \mbox{F-DENSER++} because it has been proved to be able to generate highly performing fully-trained models, in less time than the standard DENSER implementation. 

\section{Evolution of Convolutional Neural Networks}
\label{sec:evolution}

The gamma-ray detector, as described in Section~\ref{sec:simulation}, is composed by 3m $\times$ 1.5m individual stations that occupy a full circle array with a radius of approximately 80m. Therefore, each event is a matrix with the recorded signal by each of the cells. The goal is to, based on the signal matrix, distinguish between gamma radiations and protons. CNNs~\cite{lecun1998gradient} are suit for analysing spatially-correlated data, and thus appropriate for this supervised classification task. 

CNNs are a Deep Learning (DL) model, i.e., from the raw data (i.e., the matrix of signal), the model designs the features, and then performs classification based on the acquired data representation. The typical structure of CNNs divides the hidden-layers in two major blocks: (i) a set of layers responsible for representation learning and feature extraction, which is formed by Convolutional and Pooling layers; and (ii) a set of layers for classification, where fully-connected layers are used (see Figure~\ref{fig:cnns}). Convolutional layers are composed by a set of learnable filters that are convolved with the input; each filter connects locally (to what is known as receptive field) to the input and is activated by different patterns, thus encoding a different feature. Pooling layers down-sample the input by aggregating neurons, and consequently reduce the number of trainable parameters. Fully-connected layers densely connect to all neurons of the input layer. 

The design of CNNs requires the definition of: (i) the topology, i.e., the number of layers, type, sequencing, and parameterisation; and (ii) the learning strategy, i.e., the learning algorithm, and its parameterisation. Instead of hand-designing a CNN that is able to solve our gamma-ray detection problem we use F-DENSER++ to automate the search.

The dataset description, the parameterisation of F-DENSER++, and the fitness function are respectively detailed in Sections~\ref{sec:dataset}, \ref{sec:exp_setup}, and \ref{sec:fitness_function}. The experimental results are presented in Section~\ref{sec:exp_results}, and are discussed in Section~\ref{sec:discussion}.

\begin{table}
    \centering
    \begin{tabular}{c|c|c}
        \textbf{Partition}  & \textbf{\#Gamma Instances} & \textbf{\#Proton Instances} \\ \hline
        Train & 22541 & 20261  \\
        Validation & 1691 & 1519 \\
        Test & 3945 & 3546  \\
        Generalisation & 13879 & 12474 \\
    \end{tabular}
    \caption{Description of the dataset partitions.}
    \label{tab:dataset}
\end{table}

\subsection{Dataset}
\label{sec:dataset}

The dataset is composed by 79856 instances (shower events) of two disjoint classes: gamma or proton. Each instance is a 100 $\times$ 45 matrix, where each position represents the energy at a specific 3m $\times$ 1.5m cell of the circular grid of radius 80m. The positions of the matrix where there are no cells (because the grid is circular and the matrix is rectangular) are set to 0.

We partition the dataset into 4 independent sets. The first 3 are used during evolution:
\begin{description}[style=multiline, labelwidth=\widthof{Validation --},
                    font=\normalfont, leftmargin=\labelwidth, align=right]
    \item[Train --] used for training the individual with the evolved learning strategy; 
    \item[Validation --] necessary for measuring the loss during the train, to perform early stopping;
    \item[Test --] applied to compute the fitness of the network after the training. This fitness value defines the quality of the individual and guides evolution.
\end{description}

The last partition is used after the end of the evolutionary search, and measures the generalisation ability of the models. If this partition was not created it would be impossible to perform an unbiased evaluation of the generated networks because evolution is conducted towards the test partition, and consequently it is expected that the networks perform well on it; that does not mean that they perform well beyond the data used during evolution. The number of instances of each partition is detailed in Table~\ref{tab:dataset}.

%proton - 0
%photon - 1
%evo-x-train: 42802 / 20261 / 22541
%evo-x-val: 3210 / 1519 / 1691
%evo-x-test: 7491 / 3546 / 3945
%x-test: 26353 / 12474 / 13879

\subsection{Experimental Setup}
\label{sec:exp_setup}

\begin{figure}[t!]
    \scriptsize
    \begin{align}
        {<}\text{features}{>} ::= & \, {<}\text{convolution}{>}  \, | \,  {<}\text{convolution}{>}\\
                   & \, | \, {<}\text{pooling}{>} \, | \,  {<}\text{pooling}{>} \\
                   & \, | \, {<}\text{dropout}{>} \, | \, {<}\text{batch-norm}{>} \\
        {<}\text{convolution}{>} ::= & \, \text{layer:conv} \, \text{[num-filters,int,1,32,256]} \\
                    & \, \text{[filter-shape,int,1,2,5]} \, \text{[stride,int,1,1,3]} \\
                    &  \, {<}\text{padding}{>} \, {<}\text{activation}{>} \, {<}\text{bias}{>}\\
        {<}\text{batch-norm}{>} ::= & \text{layer:batch-norm}\\
       {<}\text{pooling}{>} ::= & \, {<}\text{pool-type}{>} \, \text{[kernel-size,int,1,2,5]} \\
                    & \, \text{[stride,int,1,1,3]} \, {<}\text{padding}{>} \\
        {<}\text{pool-type}{>} ::= & \, \text{layer:pool-avg} \, | \, \text{layer:pool-max}\\
        {<}\text{padding}{>} ::= & \, \text{padding:same} \, | \, \text{padding:valid}\\
        {<}\text{classification}{>} ::= & \, {<}\text{fully-connected}{>} \, | \,  {<}\text{dropout}{>} \\
        {<}\text{fully-connected}{>} ::= & \, \text{layer:fc} \, {<}\text{activation}{>} \\
                    & \, \text{[num-units,int,1,128,2048} \, {<}\text{bias}{>} \\
        {<}\text{dropout}{>} ::= & \text{layer:dropput} \, \text{[rate,float,1,0,0.7]} \\
        {<}\text{activation}{>} ::= & \, \text{act:linear} \, | \, \text{act:relu} \, | \, \text{act:sigmoid}\\
        {<}\text{bias}{>} ::= & \, \text{bias:True} \, | \, \text{bias:False}\\
        {<}\text{softmax}{>} ::= & \, \text{layer:fc} \, \text{act:softmax} \, \text{num-units:2} \, \text{bias:True}\\
        {<}\text{learning}{>} ::= & \, {<}\text{bp}{>} \, {<}\text{stop}{>} \, \text{[batch\_size,int,1,50,300]} \\
                   & \, | \, {<}\text{rmsprop}{>} \, {<}\text{stop}{>} \, \text{[batch\_size,int,1,50,300]}  \\
                   & \, | \, {<}\text{adam}{>} \, {<}\text{stop}{>} \, \text{[batch\_size,int,1,50,300]} \\
         {<}\text{bp}{>} ::= & \, \text{learning:gradient-descent} \, \text{[lr,float,1,0.0001,0.1]}  \\
                   & \, \text{[momentum,float,1,0.68,0.99]} \\
                   & \, \text{[decay,float,1,0.000001,0.001]} \, {<}\text{nesterov}{>} \\
         {<}\text{nesterov}{>} ::= & \, \text{nesterov:True} \, | \, \text{nesterov:False} \\
         {<}\text{adam}{>} ::= & \, \text{learning:adam} \, \text{[lr,float,1,0.0001,0.1]} \\
                   & \, \text{[beta1,float,1,0.5,1]} \, \text{[beta2,float,1,0.5,1]} \\
                   & \, \text{[decay,float,1,0.000001,0.001]} \\
         {<}\text{rmsprop}{>} ::= & \, \text{learning:rmsprop} \, \text{[lr,float,1,0.0001,0.1]} \\
                   & \, \text{[rho,float,1,0.5,1]} \, \text{[decay,float,1,0.000001,0.001]} \\
         {<}\text{stop}{>} ::= & \, \text{[early\_stop,int,1,5,20]}
    \end{align}

    \small
    \caption{Grammar used by F-DENSER++ for the evolution of CNNs to classify between gamma and proton.}
    \label{fig:grammar}
\end{figure}

To apply F-DENSER++ to the evolution of CNNs first of all we need to define the outer-level structure and the inner-level grammar. We use the outer-level structure: [(features, 1, 30), (classification, 1, 10), (softmax, 1, 1), (learning, 1, 1)], and the grammar of Figure~\ref{fig:grammar}. The search space encompasses CNNs with between 3 and 41 layers, and all parameters including the learning strategy are encoded in the grammar.

F-DENSER++ parameters are summarised in Table~\ref{tab:exp_setup}. The table is divided into two independent sections: (i) evolutionary parameters -- specify the evolutionary engine properties (number of generations, mutation rates, etc.); and (ii) train parameters -- enumerate the learning parameters that are fixed for all networks. The default training time is of 10 minutes, and can increase in multiples by mutation.

No data augmentation strategy is used, and the dataset is pre-processed by feature-wise centering and standard deviation normalization.

\subsection{Fitness Function}
\label{sec:fitness_function}

To evaluate the fitness of each individual, we evaluate the model in the test partition, and compute the true positive rate (TPR) and false positive rate (FPR) to build the Receiver Operating Characteristic (ROC) curve; we consider the positive class as the instances classified as proton. The fitness of each individual (ind) is calculated as: 
\begin{equation*}
    \text{fitness(ind)} = max \Bigg(\frac{\text{TPR(x)}}{\sqrt{\text{FPR(x)}}}\Bigg),
    \label{eq:fitness}
\end{equation*}
where TPR(x) and FPR(x) represent the TPR and FPR of the model at the point $x$ of the FPR threshold, respectively. Since we are maximising, the models assigned with higher fitness values are those with a higher respose of TPR for each FPR point, with emphasis to points with low FPR threshold.

\begin{table}[t!]
    \centering
    \caption{Experimental parameters.}
    \label{tab:exp_setup}
    \begin{tabular}{c | c }
        \textbf{Evolutionary Parameter} & \textbf{Value}\\ \hline
        Number of runs & 10 \\ 
        Number of generations & 100\\  
        $\lambda$ & 4 \\
        Add layer rate & 25\% \\
        Duplicate layer rate & 15\% \\
        Remove layer rate & 25\% \\
        DSGE-level rate & 15\% \\ 
        Train time rate & 20\% \\
        & \\
        \textbf{Train Parameter} & \textbf{Value} \\ \hline
        Default train time & 10 minutes \\
        Loss & Categorical Cross-entropy \\
    \end{tabular}
\end{table}

The choice of the fitness function is connected with the fact that the observation of astrophysical gamma-ray sources relies on the identification of gamma-rays which are immersed in a huge cosmic ray (hadronic) background. As the background is continuous and isotropic, while gamma-ray are localized in space, if one acquires during enough time, an excess of events coming from the gamma-ray sky region should be visible. To state that there is an excess, the number of gamma-ray events has to be greater than the fluctuations of the background. As events are considered independent the fluctuations follow the Poisson distribution, i.e., the square root of the number of events measured. By taking the number of background events much greater than the number of signal events, one can neglect the signal contribution in the square root which finally leads to the chosen fitness equation.

\subsection{Experimental Results}
\label{sec:exp_results}

\begin{figure}
    \centering
    \includegraphics[width=0.49\textwidth]{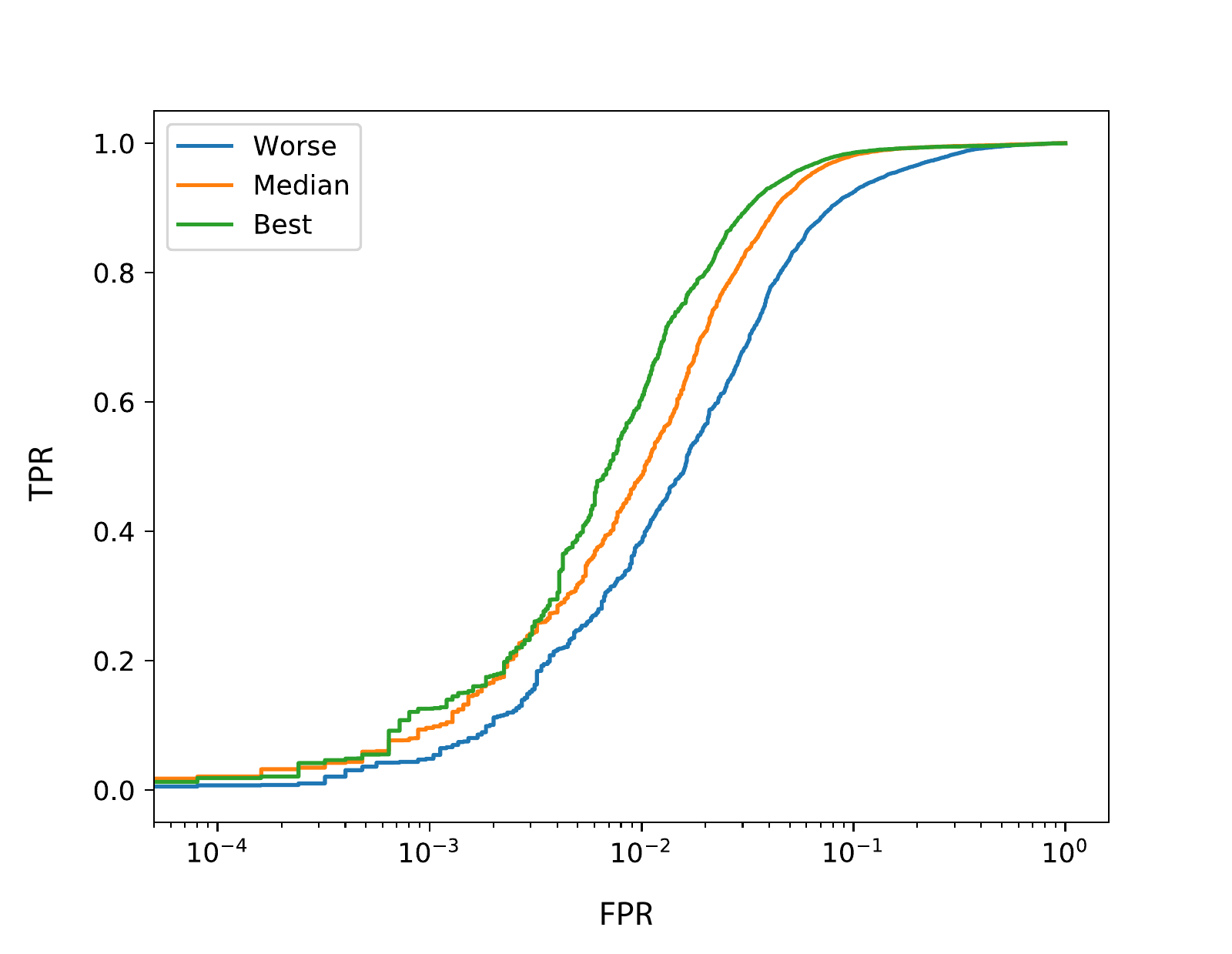}
    \caption{ROC curves of the worse, median, and best fittest individuals. A logarithmic scale is used.}
    \label{fig:average_curve}
\end{figure}

The analysis of the experimental results focuses on the performance of the evolved networks, measured on the evolutionary test set. The fitness function described in Section~\ref{sec:fitness_function} is strictly related to the ROC curve, and thus in Figure~\ref{fig:average_curve} we depict the ROC curves (measure over the generalisation set) of the fittest networks that achieve the worse, median, and highest fitness values. The fittest networks are selected according to their fitness value on the test set. 

The curve of the individual with the median fitness value is close to the curve of the best individual, indicating that the results are consistent, i.e., a high performing network is not discovered by change, but is rather an outcome of the evolutionary search of F-DENSER++. The minimum, average, and maximum fitness values are 4.07, 5.27, and 6.26, respectively.

Despite the importance of the analysis of the overall results, the ultimate goal is to select a model that is capable of addressing the problem we have at hand, in this case, a CNN which is capable of classifying between gamma and proton. We select the best performing network according to the evolutionary test fitness. Recall that this choice is not biased because we will be later comparing the results based on a different, disjoint, set of instances.

\begin{figure}
    \centering
    \includegraphics[width=0.38\textwidth]{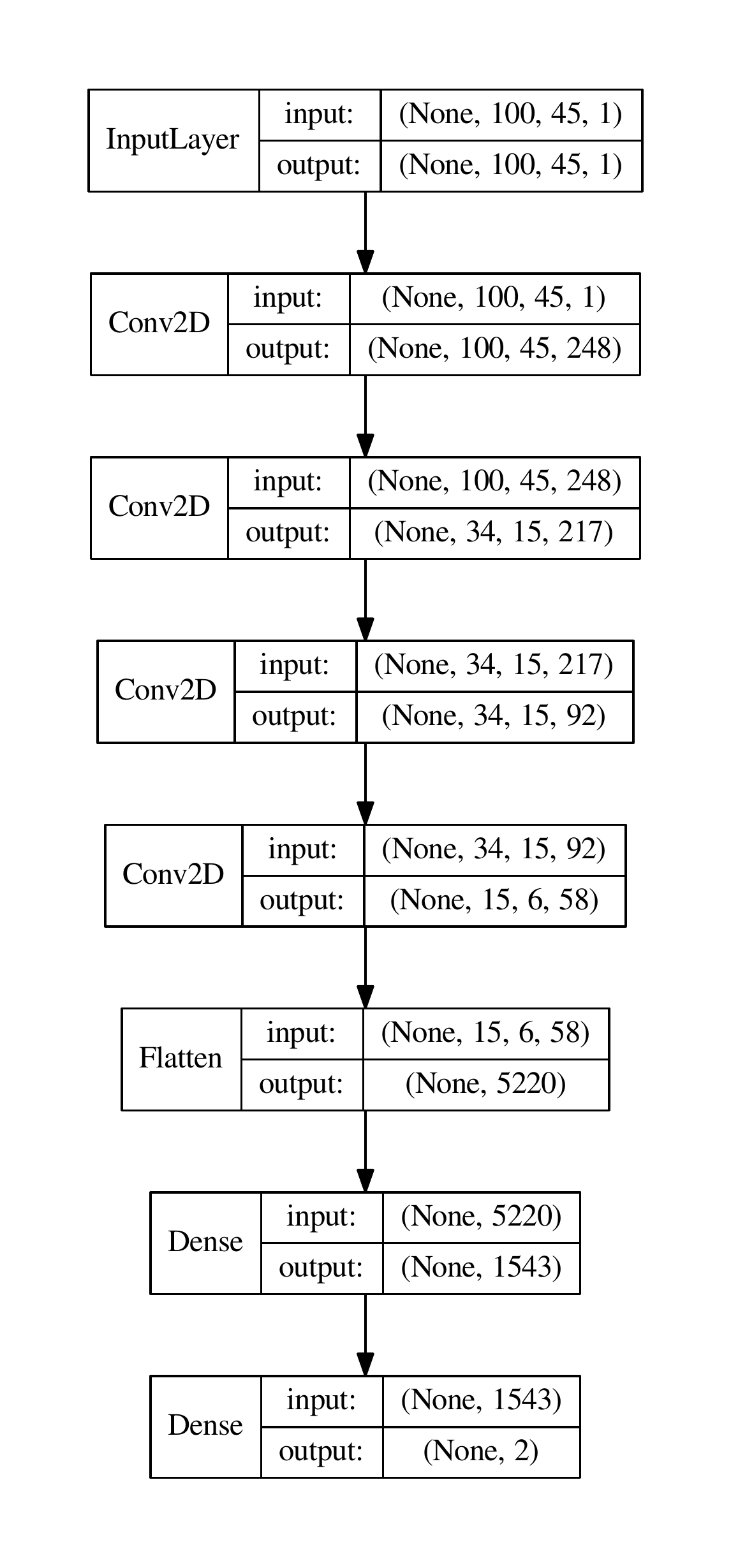}
    \caption{Topology of the fittest CNN discovered by F-DENSER++.}
    \label{fig:network_topology}
\end{figure}

The topology of the best performing network is shown in Figure~\ref{fig:network_topology}. The CNN is composed by 5 hidden-layers: 4 convolutional, and 1 fully-connected; contrary to what is common in hand-designed CNNs there are no pooling layers which demonstrates that evolution helps generating novel and out of the box topologies that human-designers would hardly think of. The fittest CNN is trained using the Adam~\cite{kingma2014adam} learning algorithm with a learning rate of $0.0001$, a beta 1 of $0.75192$, a beta 2 of $0.91021$; the learning rate decay is $0.00047$, and the batch size is 98. The fittest CNN is compared with the performance obtained by other approaches in Section~\ref{sec:discussion}.

Given the applicational type of the task, although it is not required for the network to perform in critical-time it is important that it predicts fast. The network reports an average prediction time of approximately 109 ms, i.e., 10 frames-per-second; this time includes the pre-processing. To enable future comparisons it is important to mention that the experiments are conducted in a dedicated machine, with 4 NVIDIA 1080 Ti GPUs (each with 12GB), 64 GB of RAM, and a Intel Core i7-6850K @ 3.60GHz CPU. The predictions are carried out at the CPU level; GPUs are used for training.

\subsection{Discussion}
\label{sec:discussion}

\begin{figure}[t!]
    \centering
    \includegraphics[width=0.49\textwidth]{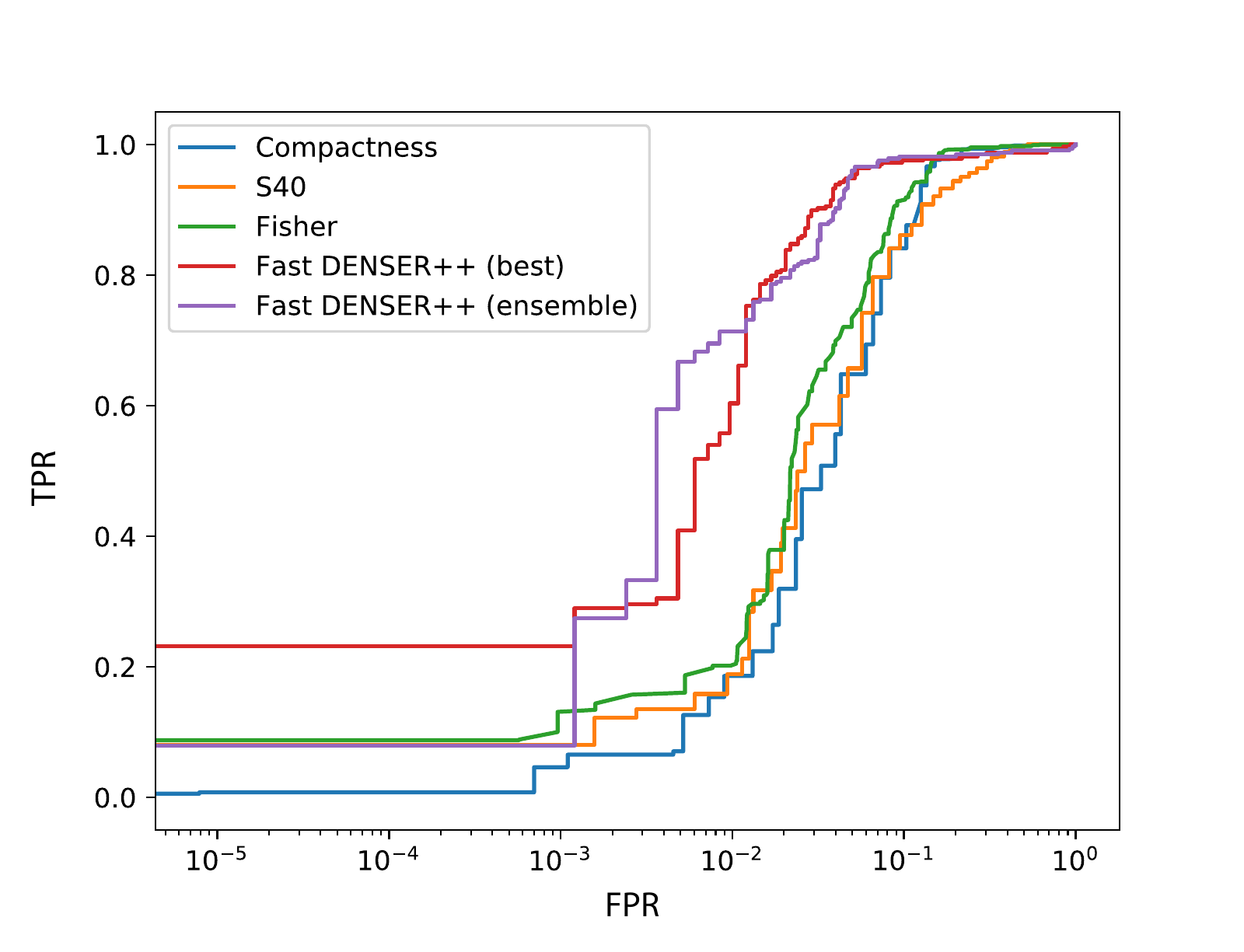}
    \caption{Comparison between the CNNs discovered by F-DENSER++ (best and ensemble) and other ML methods: Compactness, S40, and Fisher. A logarithmic scale is used.}
    \label{fig:comparison_roc}
\end{figure}

Figure~\ref{fig:comparison_roc} compares the ROC curves of the fittest CNN discovered by F-DENSER++, with the performance reported by the classic statistics (Compactness, S40, and Fisher). In addition to the fittest network we also investigate the performance of the ensemble formed by the best networks (one from each run); the generated networks are diverse in topology, and consequently are likely to be better suit for some patterns of inputs over others, i.e., while some of the networks can fail to predict a specific instance others can predict it correctly. The ensemble is formed by 10 voters (the fittest CNN of each evolutionary run), and the predicted class is computed based on the maximum of the average confidences. For all methods the performance is measured on the same partition of the data, and thus the results are comparable. The data is the same of \cite{LATTES}, but distinct from the one used in the evolutionary experiments; nonetheless, it generated from the same source. It consists of 1158 instances: 328 gamma, and 830 protons. The dataset is unbalanced and follows the distribution expected in nature.

The analysis of the plot shows that the CNNs generated by F-DENSER++ surpass the results obtained by the classical statistics. Further, the fitness values of the fittest CNN, ensemble, Compactness, S40, and Fisher are of approximately 8.34, 9.89, 3.13, 3.35, and 4.22, respectively. Comparing to the best result of the classic statistics, the generated CNNs promote improvements by a factor of 2, and 2.3 for the fittest CNN, and ensemble, respectively. 

\section{Conclusions and Future Work}

Gamma-ray detection helps investigating extreme phenomena in the Universe, e.g., gamma-ray burst arising from fast rotating neutron stars or supermassive black holes. In this work it is our objective to use deep learning to improve the gamma/hadron discrimination, based on the patterns they produce at ground impact. The impact patterns are stored as matrices of signal, where each position keeps the energy detected in a specific WCD. Therefore, this task is suitable for the application of CNNs: a deep learning network that is known for its ability to learn to distinguish patterns in complex signals.

The problem associated to the deployment of CNNs is related to the design and parameterisation difficulties: the networks are composed by several layers, each with specific parameters; in addition to the definition of the layers and their sequence we require the choice for the most effective learning algorithm and its hyper-parameters. To overcome this challenge we use F-DENSER++ to automatically search for an effective CNN for our gamma-ray detection problem. 

The results show that not only is the CNN generated by F-DENSER++ able to solve the gamma-ray detection problem, but it does so surpassing the performance reported by previous \emph{classic} methods, namely \emph{compactness}, \emph{S40}, and \emph{Fisher}. Whilst the fittest CNN reports a fitness value of 8.34, the best performance of the classic methods is of 4.22, i.e., an improvement by a factor of 2. This result is even more surprising when forming an ensemble composed by the 10 best CNNs: the fitness increases from 8.34 to 9.89, that corresponds to an improvement by a factor of 2.3. 

Future work will expand in two separate directions: (i) investigate the performance of F-DENSER++ in the search for CNNs for different primary energy; and (ii) study the impact of the detector configuration on the detection performance (e.g., number and shape/dimensions of the sensors). In terms of evolution we will incorporate the number of layers and trainable parameters in the evolutionary objectives, with the rationale of generating more compact networks that may be easier to analyse and validate; this will be carried in a multi-objective fashion.

\section*{Acknowledgements}
\noindent This work is partially funded by: Funda\c{c}\~ao para a Ci\^encia e Tecnologia (FCT), Portugal, under the grant SFRH/BD/114865/2016.

\bibliographystyle{IEEEtran}  
\bibliography{bibliography}

%\EOD

\end{document}